\documentclass[journal]{IEEEtran}
\usepackage{amsmath,amsfonts}
\usepackage{array}
\usepackage{textcomp}
\usepackage{stfloats}
\usepackage{url}
\usepackage{verbatim}
\usepackage{graphicx,subcaption}
\usepackage{cite,color}
\usepackage[switch]{lineno}
\usepackage{algorithm}
\usepackage[noend]{algpseudocode}
\begin{document}
\title{Distributed Multi-Head Learning Systems for Power Consumption Prediction}

\author{Jia-Hao~Syu, Jerry~Chun-Wei~Lin$^{*}$, Philip S. Yu
\thanks{Jia-Hao Syu is with the Department of Computer Science and Information Engineering, National Taiwan University, Taiwan. Email: f08922011@ntu.edu.tw}
\thanks{Jerry Chun-Wei Lin is with the Department of Computer Science, Electrical Engineering and Mathematical Sciences, Western Norway University of Applied Sciences, Bergen, Norway. Email: jerrylin@ieee.org (*Corresponding author)}
\thanks{Philip S. Yu is with the Department of Computer Science, University of Illinois at Chicago, USA. Email: psyu@cs.uic.edu}
}

\markboth{}%
{}

% \IEEEpubid{0000--0000/00\$00.00~\copyright~2022 IEEE}

\maketitle

\begin{abstract}
As more and more automatic vehicles, power consumption prediction becomes a vital issue for task scheduling and energy management.
Most research focuses on automatic vehicles in transportation, but few focus on automatic ground vehicles (AGVs) in smart factories, which face complex environments and generate large amounts of data.
There is an inevitable trade-off between feature diversity and interference.
In this paper, we propose Distributed Multi-Head learning (DMH) systems for power consumption prediction in smart factories.
Multi-head learning mechanisms are proposed in DMH to reduce noise interference and improve accuracy.
Additionally, DMH systems are designed as distributed and split learning, reducing the client-to-server transmission cost, sharing knowledge without sharing local data and models, and enhancing the privacy and security levels.
Experimental results show that the proposed DMH systems rank in the top-2 on most datasets and scenarios.
DMH-E system reduces the error of the state-of-the-art systems by 14.5\% to 24.0\%.
Effectiveness studies demonstrate the effectiveness of Pearson correlation-based feature engineering, and feature grouping with the proposed multi-head learning further enhances prediction performance.
\end{abstract}


% OK ->
\section{Introduction}
\IEEEPARstart{I}{ndustrial} revolution started with the steam engine in the 18th century.
The second industrial revolution belongs to the mass production and assembly lines in the early 20th century, and the third industrial revolution credits to the computer control systems for production in the late 20th century~\cite{BigDataSurvey}.
Over the past decade, the fourth industrial revolution, also known as intelligent manufacturing, has been widely discussed.
When it comes to intelligent manufacturing, production emphasizes customization, flexibility, agility, and precision~\cite{IM}.
To reduce the costs of frequently changing production, production chains of smart factories employ numerous robot stations and automatic vehicles, which are power-sensitive and environment-sensitive due to extensive collaboration on production chains.
Therefore, power consumption prediction becomes a vital issue in smart factories, and acts as the decision-making basis of charging scheduling~\cite{Bus,EV}, task scheduling~\cite{TSF1,New-4}, energy management~\cite{M-IoT,M-BDQ,M-SUS}, and anomaly detection~\cite{SMC}.

Artificial intelligence technologies are broadly used for power consumption prediction~\cite{SF}.
Internet of Things (IoT) sensors extract real-time data; data mining explores meaningful patterns in the collected data, and machine learning makes further predictions and arrangements.
However, most research on power consumption prediction focuses on smart cities~\cite{ENG} and smart transportation~\cite{Bus,EV}, but little attention has been paid to smart factories and automatic ground vehicles (AGVs).
% Gao et al.~\cite{Bus} optimized charging scheduling for electric buses through power consumption prediction; in contrast, Almaghrebi et al.~\cite{EV} developed charging demand prediction models to schedule electric vehicle charging.
% Javid et al.~\cite{AD-AGV} detected unexpected power consumption in automated vehicles via deep learning, and Benecki et al.~\cite{SMC} predicted the power consumption of automatic grounded vehicles in a smart factory.
% Most research on power consumption prediction focuses on electric and automatic vehicles in transportation, but little attention has been paid to automatic ground vehicles (AGVs) in smart factories.

AGVs in smart factories face complex environments and various tasks.
Massive information collected by sensors may contain noise, unstable transmission quality, and limited connection bandwidth to the management system.
However, accurate prediction of power consumption relies heavily on the collection and analysis of high-quality data.
Traditional prediction systems feed all data directly into the model, but this leads to overfitting and noise, and there is an inevitable trade-off between feature richness and noise~\cite{SMC}.
The organization and fusion of the collected features are critical for both temporal and spatial predictions~\cite{New-5}.
Furthermore, exchanging knowledge and experience (decision process embedded in model parameters) learned from AGVs is also beneficial for prediction.
Efficient but secure knowledge sharing between AGVs is an open and emerging research problem to alleviate concerns of production and commercial confidentiality (between smart factories).
In this paper, we focus on predicting power consumption of AGV using time-series features, and focus on real-time communication through distributed learning algorithms.

To overcome the above-mentioned issues, we propose Distributed Multi-Head learning (DMH) systems for power consumption prediction, designed for real-world implementation in smart factories.
To make full use of collected features, the proposed DMH first performs feature engineering, and groups features according to Pearson's correlation, and adopts the rolling window mechanism to pack the time-series features.
We then propose multi-head learning mechanisms that obtain multiple head networks for feature groups (one head network per feature group) to reduce noise interference.
The head network is designed to predict the corresponding features at the next time step, which are aggregated as input to the prediction network to predict power consumption.
A derivative system is also proposed to enable the head networks to directly predict power consumptions, which are aggregated as input to make the final prediction of power consumption.
The derivative system is called the ensemble DMH system, DMH-E; the original system is called the time-series DMH system, DMH-T.
Since DMH systems are multi-task learning, we design a loss balancing mechanism to balance the scale and variance of different losses.

Moreover, the proposed DMH systems are designed for distributed learning, more precisely for split learning. The head networks reside on the client (AGV), and the prediction network resides on the server (prediction system).
DMH systems do not share local data (features) and transmit only the predicted features to the server, which not only reduces the transmission cost but also increases privacy and security.
The prediction network located on the server can aggregate the knowledge of clients and help update the main network using the transmitted gradients, which share the knowledge without sharing local data and the whole model.
The contributions of this paper can be summarized as follows:
\begin{enumerate}
\item The designed feature engineering and proposed multi-head learning improve the accuracy of power consumption predictions.
\item Designed split learning improves the privacy and security levels of DMH and shares knowledge without sharing local data and the entire model.
\item Distributed learning developed in DMH reduces the client-to-server transmission cost to $\frac{1}{W}$, where $W$ is the feature window size.
\end{enumerate}

Experimental results show that the proposed DMH systems rank in the top-2 on most datasets and scenarios.
By comparison, DMH-E generally outperforms DMH-T and reduces the mean absolute error (MAE) of the state-of-the-art system~\cite{SMC} by 14.5\% to 24.0\%.
For robustness evaluation, we conduct the systems to predict power consumption after 5 and 10 time steps.
The proposed DMH-E still outperforms the state-of-the-art system~\cite{SMC} by 8.9\% to 53.1\% (after 5 time steps) and 2.0\% to 58.3\% (after 10 time steps).
Effectiveness studies first demonstrate the effectiveness of Pearson's correlation-based feature engineering, and feature grouping with the proposed multi-head learning further enhances prediction performance.

The following sections are organized as follows.
Section~\ref{sec:LR} surveys related works on power consumption prediction, feature engineering, and distributed learning.
The proposed DMH systems are presented in Section~\ref{sec:Sys}, including an introduction to the designed feature engineering, distributed multi-head learning, and loss balancing mechanism.
Section~\ref{sec:ES} states the environment setup for the following experiments, and Section~\ref{sec:Exp} presents and discusses the experimental results, and we summarize the paper in Section~\ref{sec:Con}.
% OK <-

% OK ->
% ====================
\section{Literature Review}   \label{sec:LR}
In this section, we survey the related works of power consumption prediction in Section~\ref{sec:LR-PCP}, and the feature engineering in Section~\ref{sec:LR-FS}.
The distributed learning of federated and split learnings are introduced in Section~\ref{sec:LR-DL}.

% OK ->
\subsection{Power Consumption Prediction}   \label{sec:LR-PCP}
Power consumption prediction is vital to the fourth industrial revolution, and acts as the decision-making basis for smart factories.
Smart factories face complex environments and tasks, where sensor data can be noisy, have unstable transmission quality, and limited bandwidth~\cite{Sensor}.
Traditional approaches use all collected data, risking overfitting and noise.
However, accurate power consumption prediction depends on the high-quality and effective usage of data.
Thus, balancing feature richness and noise is a crucial research question~\cite{SMC}, and the fusion of collected features is essential for both temporal and spatial prediction~\cite{New-5}, especially in power consumption prediction.
For example, Li et al.~\cite{MVDNN} predicted the fuel consumption of vehicles based on multi-view deep neural network, and Benecki et al.~\cite{SMC} predicted the power consumption of automatic vehicles in a smart factory.
Both studies used neural networks to predict vehicle power consumption, and focused on the selection and utilization of multiple features.

The prediction of power consumption can assist in charging scheduling, task scheduling, and anomaly detection.
From the perspective of vehicles, Gao et al.~\cite{Bus} optimized the charging scheduling of electric buses through power consumption prediction on a wavelet neural network.
From the perspective of charging stations, Almaghrebi et al.~\cite{EV} developed charging demand prediction models to further manage and arrange the charging of electric vehicles.

After predicting or observing power consumption, the information can assist in task scheduling, especially in cloud services and smart factories.
Sathya and GaneshKumar~\cite{TS} optimized task scheduling of cloud services to minimize energy consumption; Tang et al.~\cite{TSDC1} and Naik et al.~\cite{TSDC2} developed energy-aware task scheduling systems in data centers.
Nouri et al.~\cite{TSF1} and Faria et al.~\cite{TSF1} proposed manufacturing and process scheduling systems with minimized power consumption and production costs.
The above literature demonstrates the importance of power consumption prediction and task scheduling for minimizing power consumption.

In addition to task scheduling, power consumption prediction can also assist in anomaly detection.
Benecki et al.~\cite{SMC} predicted the power consumption of automatic vehicles in a smart factory, and the difference between predicted and actual power consumption could provide information for anomaly detection (large differences may be caused by abnormal events).
% Javid et al.~\cite{AD-AGV} detected unexpected power consumption in automated vehicles via ensembled deep neural networks with attention layers.
The concept can be extended to predictive maintenance, which provides early warning of potential anomalies~\cite{PM}.
% OK <-

% OK ->
\subsection{Feature Engineering}   \label{sec:LR-FS}
Feature engineering attempts to systematically extract features from raw data, including feature selection and transformation~\cite{FE}.
Feature selection aims to analyze, rank, select relevant, and remove redundant features~\cite{FS}.
Statistical techniques are frequently used for feature selection, such as correlation coefficients, p-values, z-tests, and t-tests.
Filter-based feature selection methods remove features by thresholding statistical values.
Benecki et al.~\cite{SMC} selected features for AGV power consumption prediction based on Pearson's correlation coefficient, demonstrating the effectiveness of statistics-based feature selection.
However, feature selection is based on the distribution and domain knowledge of the dataset, and lacks generality.

On the other hand, some models do not select features, but transform and arrange all features.
Neural network techniques of Convolutional Neural Network (CNN), Recurrent Neural Network (RNN), and autoencoder are commonly used for feature transformation and extraction~\cite{FT}.
In order to take full advantage of features, we adopt all features in this paper, and organize the used features through static analysis, and then perform feature transformation through various neural networks.
% OK <-

% OK ->
\subsection{Distributed Learning}   \label{sec:LR-DL}
With the increasing discussion of edge computing in intelligent manufacturing, distributed learning~\cite{DL} has become an emerging research topic in edge computing~\cite{New-1}, wireless networks~\cite{wirless}, and image processing~\cite{New-2, New-3}.
Distributed learning is known for decentralized and parallelized learning techniques, which distribute learning tasks to multiple nodes (edges) to achieve parallelization and scalability, and is suitable for big data and IoT environments such as smart cities and smart factories.
Distributed learning algorithms face challenges of efficiency~\cite{DLE}, load balancing, transmission bandwidth~\cite{DLC}, and privacy~\cite{DLS}.
Balancing the above challenges is an emerging research topic recently, and has led to two main branches of algorithms, federated learning, and split learning.

Federated learning aims to collaboratively train models using data from multiple clients without sharing data~\cite{FL}, and is decentralized, assigning learning tasks to clients, aggregating trained models, and broadcasting the last version.
Contrary to centralized learning algorithms, where all data are collected and trained on a powerful server, federated learning distributes the last model to each client, and clients independently train the models using local (own) data, and send the trained models back to the server.
Federated learning shares knowledge (model) but not local data, and is a privacy-aware distributed learning, and widely discussed with IoT~\cite{FLI}, communication~\cite{FLM}, healthcare~\cite{FLH}, and smart factory~\cite{FLSF}.
However, federated learning still suffers from various attacks~\cite{FLA}.
For example, if the algorithm suffers from a man-in-the-middle attack, each client's initial and trained models will be snooped.
Hackers can obtain local data by reverse engineering each client's trained model.

To further improve the privacy and security levels of federated learning, numerous split learning algorithms have been developed~\cite{SL}.
Split learning aims to train a model collaboratively by multiple clients without sharing data and the entire model, which is split between the client and the server.
A basic split learning algorithm splits the model into two parts (the client holds the first part, and the server holds the second part), including four steps, as described below.
\begin{enumerate}
\item The client inputs the local data to the first part of the model, and forward propagates until the end, and sends the output to the server.
\item The server continues the forward propagation until the end, and calculates the loss.
\item The server backward propagates until the head, and obtains the gradients to update the second part of the model, and sends the last gradient to the client.
\item The client continues the backward propagation to update the first part of the model.
\end{enumerate}
Apart from high privacy and security levels, split learning also has advantages in terms of communication efficiency (no need to send the whole model, just send the weights and gradients of one layer)~\cite{SLE} and low computing overhead on clients (edge devices).
% OK <-
% OK <-

\begin{figure}[!t] 
  \centering
  \includegraphics[width=3.25in]{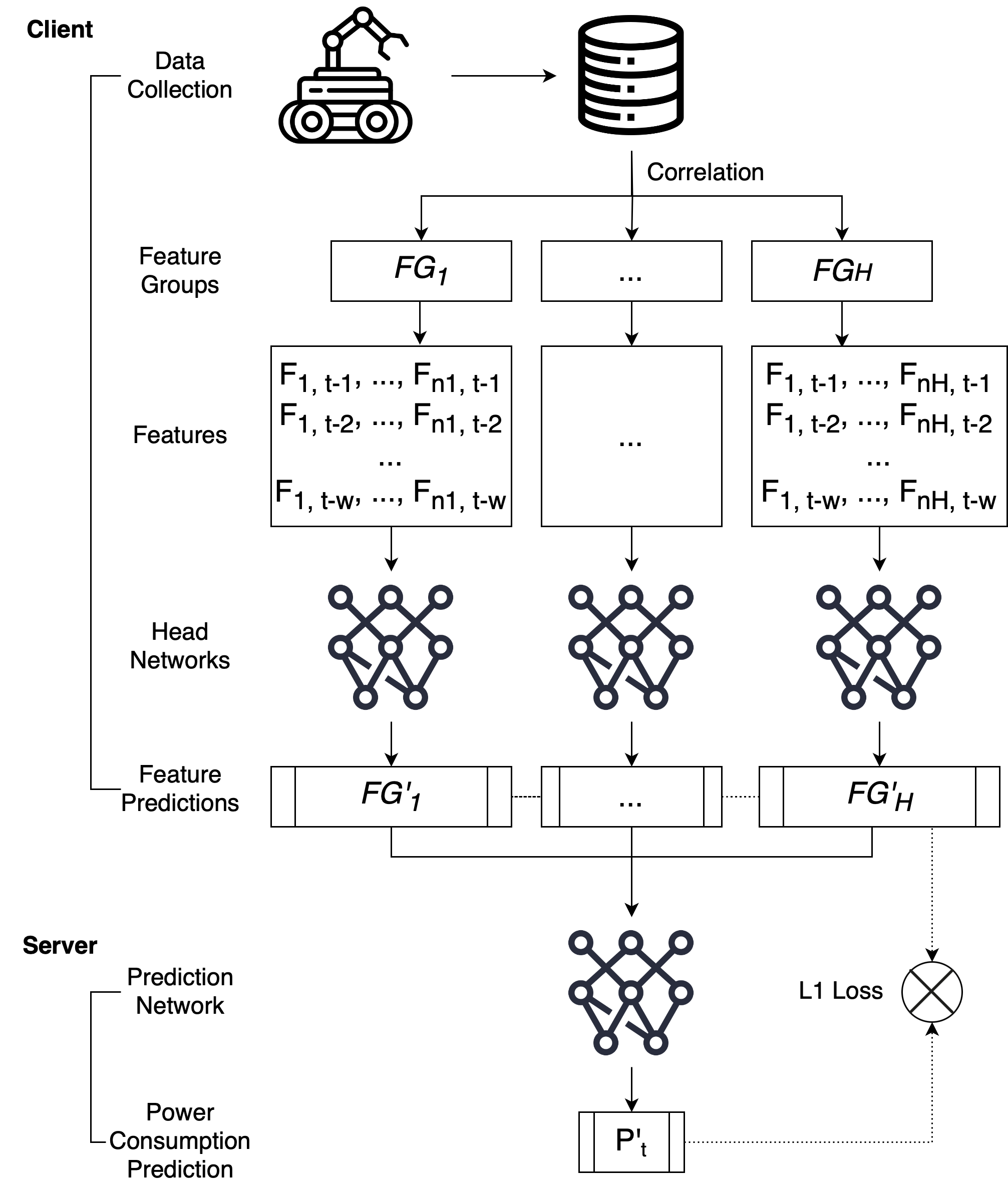}
  \caption{Flow chart of the proposed DMH}
  \label{fig:FR}
\end{figure}

% OK ->
% ====================
\section{Proposed DMH Systems}   \label{sec:Sys}
In this paper, the research task is predicting real-time power consumption of AGV using time-series features.
The proposed DMH systems are presented in this section, and Fig.~\ref{fig:FR} illustrates the flow chart of the DMH.
We first introduce the feature engineering that groups features, as described in Section~\ref{sec:Sys-FE}.
The distributed multi-head learning is then presented in Section~\ref{sec:Sys-DMH}, including the distributed mechanism in Section~\ref{sec:Sys-DMH-DL}, and the design of head networks in Section~\ref{sec:Sys-DMH-HN}, and the loss balancing mechanism in Section~\ref{sec:Sys-DMH-LB}.
Finally, Section~\ref{sec:Sys-PC} provides a pseudo-code for overviewing the proposed systems.
For readability, we list all abbreviations and variables of the proposed DMH in Table~\ref{tab:ADD}.

\begin{table}[t]
\caption{Abbreviations and Variables of DMH}
\centering
\begin{tabular}{|l|l|}
\hline
Abbreviation & Description \\
\hline
AGV   & Automatic Ground Vehicle \\
DMH   & Distributed Multi-Head learning \\
DMH-E & Ensemble DMH \\
DMH-T & Time-series DMH \\
MAE   & Mean Absolute Error \\
MSE   & Mean Square Error \\
\hline
\multicolumn{2}{c}{} \\
\hline
Variable & Description \\
\hline
$\text{F}_{i}$ & feature, $i=1, \ldots, M$ \\
$FG_{h}$       & feature group, $h=1, \ldots, H$ \\
$FG^{'}_{h,t}$ & predicted feature group $h$ at time $t$ by DMH-T \\
$W$            & window size of the features \\
$M$       & numbers of features \\
$H$       & numbers of feature groups \\
$\text{P}_{t}$ & power consumption at time $t$ \\
$\text{P}^{'}_{h, t}$ & preliminary prediction of power consumption by the \\
& head network $h$ of DMH-E at time $t$ \\
$\text{P}^{'}_{t}$    & predicted power consumption by DMH \\
$C_{i}$   & Pearson's correlation coefficients between $F_{i}$ and P \\
$T_{i}$   & coefficient thresholds, $i=1, \ldots, H+1$ \\
$n_{h}$   & number of features in $FG_{h}$ \\
$L_{h,i}$ & loss of the head $h$ at epoch $i$ \\
$M_{h,i}$ & dynamic loss multiplier of the loss $h$ at epoch $i$ \\
\hline
\end{tabular}
\label{tab:ADD}
\end{table}

% OK ->
\subsection{Feature Engineering}   \label{sec:Sys-FE}
Unlike literature~\cite{SMC} that ignores low correlation features, we design a grouping mechanism based on Pearson's correlation in feature engineering.
Suppose there are $M$ features ($\text{F}_{1}, \ldots, \text{F}_{M}$) divided into $H$ groups ($FG_{1}, \ldots, FG_{H}$), where each feature is a time series.
We first compute Pearson's correlation coefficients between each feature and power consumption (P) in the training data, denoted as $C_{1}, \ldots, C_{M}$.
$C_{m}$ is defined as
\begin{equation}     \label{eq:cor}
C_{m} = \frac{E[~ (\text{F}_{m}-\mu_{\text{F}_{m}})~(\text{P}-\mu_{\text{P}}) ~]}{\sigma_{\text{F}_{m}}~\sigma_{\text{P}}},
\end{equation}
where $\mu_{\text{F}_{m}}$ and $\sigma_{\text{F}_{m}}$ are the mean and standard deviation of $\text{F}_{m}$, and $\mu_{\text{P}}$ and $\sigma_{\text{P}}$ are the mean and standard deviation of P.
For the $H$ group, there are $H+1$ coefficient thresholds $T_{0}, \ldots, T_{H}$, where $T_{0} = 0$ and $T_{H} = 1$.
The features in the group $h$, $FG_{h}$ ($h = 1, \ldots, H$), are defined as:
\begin{equation}     \label{eq:FG}
FG_{h} = \{~\text{F}_{m} ~|~ T_{h-1} \leq C_{m} < T_{h},~ m = 1, \ldots, H~\}.
\end{equation}
Let $n_{h}$ be the number of features in group $h$, which means $n_{h} = ~|FG_{h}|$.

Furthermore, we employ rolling windows for network training with a window size of $W$.
The features for predicting time power consumption at time $t$ ($\text{P}_{t}$) include $FG_{h, t-1}, \ldots, FG_{h, t-W}$, where $h = 1 , \ldots, H$.
% OK <-

\begin{figure}[!t] 
  \centering
  \includegraphics[width=3.25in]{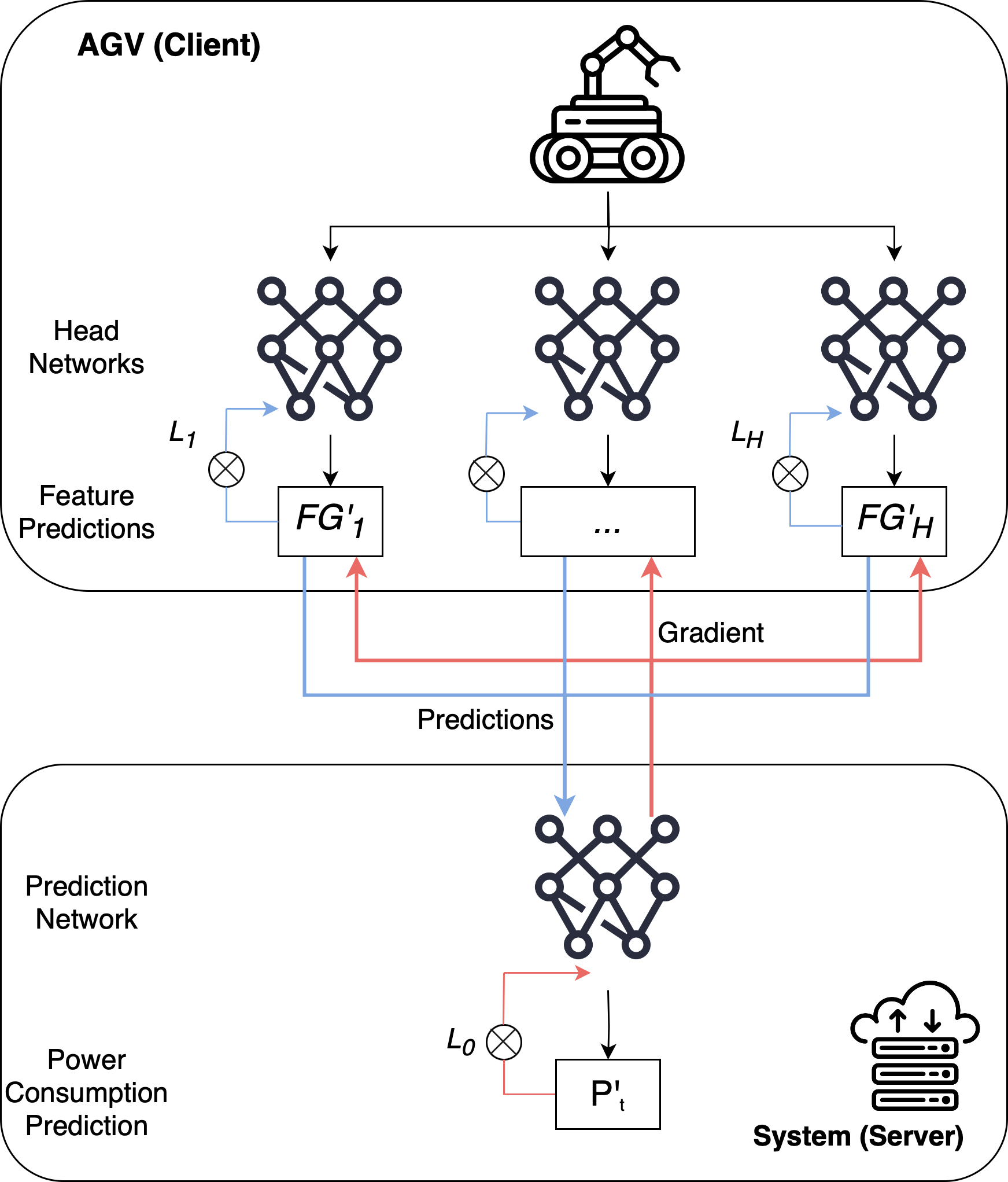}
  \caption{Designed distributed learning mechanism}
  \label{fig:DL}
\end{figure}

\begin{table*}[t]
\caption{Network Design}
\centering
\begin{tabular}{|l|l|l|l|l|}
\hline
\multicolumn{1}{|c|}{Head Network - MLP} & \multicolumn{1}{c|}{Head Network - CNN} & \multicolumn{1}{c|}{Head Network - LSTM} & \multicolumn{1}{c|}{Head Network - Transformer}        & \multicolumn{1}{c|}{Prediction Network} \\ 
% \multicolumn{2}{|c|}{- MLP}  & \multicolumn{2}{c|}{- CNN}  & \multicolumn{2}{c|}{- LSTM} & \multicolumn{2}{c|}{- Transformer} & \multicolumn{2}{c|}{Network} \\ 
% \hline
% Layer   & Neuron    & Layer   & Neuron    & Layer   & Neuron    & Layer       & Neuron    & Layer  & Neuron \\
% \hline
% Linear  & 256       & Conv1d  &           & LSTM    &           & Transformer &           & Linear &16 \\
% Sigmoid &           & Sigmoid &           & Sigmoid &           & Sigmoid     &           & LRelu  &   \\
% Linear  & 64        & Linear  & 256       & Linear  & 64        & Linear      & 16        & Linear & 4 \\
% LRelu   &           & Sigmoid &           & Sigmoid &           & Sigmoid     &           & LRelu  &   \\
% Linear  & 16        & Linear  & 64        & Linear  &$n_{h}^{*}$& Linear      & 4         & Linear & 1 \\
% LRelu   &           & LRelu   &           &         &           & Sigmoid     &           &        &   \\
% Linear  &$n_{h}^{*}$& Linear  & 16        &         &           & Linear      &$n_{h}^{*}$&        &   \\
%         &           & LRelu   &           &         &           &             &           &        &   \\
%         &           & Linear  &$n_{h}^{*}$&         &           &             &           &        &   \\
% \hline
\hline
Linear (Sigmoid, $n = 256$)       & Conv1d (Sigmoid)                  & LSTM (Sigmoid)                    & Transformer (Sigmoid)             & Linear (Sigmoid, $n = 16$) \\
Linear (Sigmoid, $n = 64$)        & Linear (Sigmoid, $n = 256$)       & Linear (Sigmoid, $n = 64$)        & Linear (Sigmoid, $n = 16$)        & Linear (Sigmoid, $n = 4$)  \\
Linear (Sigmoid, $n = 16$)        & Linear (Sigmoid, $n = 64$)        & Linear (Sigmoid, $n = n_{h}^{*}$) & Linear (Sigmoid, $n = 4$)         & Linear (Sigmoid, $n = 1$)  \\
Linear (Sigmoid, $n = n_{h}^{*}$) & Linear (Sigmoid, $n = 16$)        &                                   & Linear (Sigmoid, $n = n_{h}^{*}$) &                           \\
                                  & Linear (Sigmoid, $n = n_{h}^{*}$) &                                   & &                           \\
\hline
\end{tabular}
\label{tab:ND}
\end{table*}

% OK ->
\subsection{Distributed Multi-Head Learning}   \label{sec:Sys-DMH}
In this paper, we design a Distributed Multi-Head Learning mechanism (DMH), which takes full advantage of all features, and distributes learning tasks to clients and a server with fewer data transfers.
Two branches of DMH are designed, called DMH-E and DMH-T.
Both DMH-E and DMH-T contain $H$ head networks and a prediction network.
The head network $h$ takes inputs of $FG_{h, t-1}, \ldots, FG_{h, t-W}$, which is $n_{h} \times W$ in size.

DMH-T is a DMH consisting of time-series prediction.
DMH-T is designed to first predict the features at time $t$, and then predict the power consumption, since the features of movements and motors directly reflect the power consumption.
If the network can accurately predict the features at time $t$, then the power consumption at time $t$ can also be accurately predicted.
The DMH-T is designed as follows.
The head networks of DMH-T take the inputs to predict the corresponding features of the next time step ($FG_{h, t}$), and the predictions are denoted as $FG^{'}_{1, t}, \ldots, FG^{'}_{H, t}$.
The prediction network then takes $FG^{'}_{1, t}, \ldots, FG^{'}_{H, t}$ as input for final prediction of power consumption, $\text{P}^{'}_{t}$.
There are $H+1$ losses in DMH-T, which is the MAE.
The $H$ head losses update the corresponding head network, and the final prediction error updates the prediction and all $H$ head networks.

DMH-E is an ensemble DMH that aims to independently predict the power consumption through different feature groups with different degrees of correlation, and integrate the predictions.
The DMH-E is designed as follows.
The head networks of DMH-E take the inputs to predict the preliminary power consumption, denoted as $\text{P}^{'}_{1, t}, \ldots, \text{P}^{'}_{H, t}$.
The prediction network then integrates $\text{P}^{'}_{1, t}, \ldots, \text{P}^{'}_{H, t}$ to make the final prediction $\text{P}^{'}_{t}$.
There are also $H+1$ losses in DMH-E, which is MAE.
The $H$ head losses update the corresponding head network, and the final prediction error updates the prediction and all $H$ head networks.
% OK <-

% OK ->
\subsubsection{Distributed Learning}   \label{sec:Sys-DMH-DL}
As a distributed learning system, we split the DMH into two sub-networks, the head networks and the prediction network, referred from the basic split learning algorithm described in Section~\ref{sec:LR-DL}.
The head networks are allocated on the client (AGV), and the prediction network is allocated on the server (prediction system), as shown in Fig.~\ref{fig:DL}.

For an AGV, the head networks are responsible for predicting features at time $t$ (DMH-T) or power consumption (DMH-E).
The AGV (client) sends the predictions of the head networks to the prediction system (server), and the prediction system makes the final prediction of power consumption ($\text{P}^{'}_{t}$).
The server also calculates the loss of $\text{P}^{'}_{t}$, and backpropagates to perform gradient descent on the prediction network, and sends the gradient to the client to update the head networks.
The head networks are also updated through the corresponding prediction and loss.

In DMH, distributed learning and split learning are designed to distribute learning tasks to clients and the server, reducing transmission costs (transmits $FG^{'}_{1, t}, \ldots, FG^{'}_{H, t}$ only).
If all learning tasks are allocated to the server, the client should transmit all $FG_{h, t-i}, h=1, \ldots, H$ and $i=1, \ldots, W$, which requires a large transmission bandwidth because predictions are time-sensitive.
If all learning tasks are allocated to the clients, it may take a long time due to clients' limited computing resources.
Therefore, the designed DMH can reduce the transmission cost to $\frac{1}{W}$ or reduce the computational load of the client.
% OK <-

% OK ->
\subsubsection{Head and Prediction Networks Design}   \label{sec:Sys-DMH-HN}
In the proposed DMH, the head networks can be freely designed.
We take 4 common and popular networks to demonstrate the generality in the following experiments.
The 4 networks are Multilayer Perceptron (MLP), Convolutional Neural Network (CNN), Long-Short Term Memory (LSTM), and Transformer~\cite{Transformer}.
The MLP heads consist of 4 fully connected layers with sigmoid and leaky ReLU activation functions.
The CNN heads consist of a 1-dimension convolutional layer with input and output channels of $n_{h}$, a kernel size of 3, a stride of 1, and the same padding mechanism, and are followed by 4 fully connected layers.
The LSTM heads are referred from~\cite{SMC}, which consist of a 2-layer LSTM with an input size of $n_{h}$, a hidden size of 35, and are followed by 3 fully connected layers.
The Transformer heads consist of a source and a target sequence extractor, which are two independent fully connected layers with input and output sizes of $n_{h}$.
The source and target sequences are fed into a Transformer layer~\cite{Transformer} with $W$ heads, 1 encoder, 1 decoder, 8-dimension feedforward network, and are followed by 3 fully connected layers.

The prediction network can also be freely designed by users, and we use MLP in the following experiments.
The prediction network consists of 3 fully connected layers (with 16, 4, and 1 neurons) with leaky ReLU activation functions.
The details of the network design are shown in Table~\ref{tab:ND}, where LRelu is the leaky ReLU activation function and $n_{h}^{*}$ is the number of features in the corresponding feature group for DMH-T and 1 for DMH-E.
Note that we adjust the number of neurons slightly so that the systems on different datasets (different numbers of features) have a similar number of parameters.
% OK <-

% OK ->
\subsubsection{Loss Balancing}   \label{sec:Sys-DMH-LB}
There are multiple losses in the designed DMH, and the loss distributions are different, especially for DMH-T.
In the designed DMH-T, there are $H+1$ losses, where $H$ losses are for feature predictions, and 1 loss is for power consumption prediction.
Predictions have different loss distributions, which may cause gradients to concentrate on several large losses, thus misleading power consumption predictions.
We adopt the loss balancing mechanism of~\cite{ULB} for DMH-T, based on homoscedastic uncertainty.
The dynamic loss multipliers, $M_{0}, \ldots, M_{H}$, are determined to standardize the uncertainty and magnitude of the $H+1$ losses.

Let the MAE of the power consumption prediction of the epoch $i$ be $L_{0,i}$, and let the MAE of $H$ feature predictions of the epoch $i$ be $L_{1,i}, \ldots, L_{H, i}$.
The multipliers of the period $i$ are defined as:
\begin{equation*}
\begin{aligned}
M_{0,i} &= 1, \\
M_{h,i} &= \frac{\text{std}(L_{h, i-1})}{\text{std}(L_{0, i-1})}, ~~h=1, \ldots, H,
\end{aligned}
\end{equation*}
where $\text{std}(\cdot)$ is the standard deviation, and $i$ starts from 2.
At the first epoch, all multipliers are initialized to 1.
Additionally, we have upper and lower bounds of 0.1 and 10 for all multipliers, respectively, to prevent drastic adjustments.
% OK <-

% OK ->
\subsection{Pseudo-Code}     \label{sec:Sys-PC}
Algorithm~\ref{alg:PC} presents the pseudo-code of the proposed DMH system.
For power consumption prediction, DMH takes the inputs of time series features F, feature group thresholds $T$, feature window size $W$, and the number of head networks $H$.
DMH first groups features F by thresholds $T$ and Eqs.~\ref{eq:cor} and~\ref{eq:FG}, and obtains $H$ feature groups.
DMH then packs the data with a window size of $W$ (described in Section~\ref{sec:Sys-FE}), and feeds them to the corresponding head networks.
The $H$ head networks of DMH-T predict the corresponding features at the next time step, while the $H$ head networks of DMH-E predict the preliminary power consumption.
The prediction network aggregates the output of the $H$ head networks to make the final power consumption prediction.
Note that DMH-T's prediction network collects $M$ values (the number of features) from the $H$ head networks, while DMH-E's prediction network collects $H$ values (preliminary prediction) from the $H$ head networks.

\begin{algorithm}[t]
\footnotesize 
  \caption{DMH systems}
  \label{alg:PC}
  \begin{algorithmic}[1]
    \renewcommand{\algorithmicrequire}{\textbf{Input:}}
    \Require $\text{F}$, $T$, $W$, $H$
    \State Grouping features F by Eqs.~\ref{eq:cor} and~\ref{eq:FG} with thresholds $T$
    \State Packing data $FG$ with a window size $W$
    \State Feed each $FG$ ($H$ in total) into the corresponding head neatwork
    \State Aggregate outputs of head networks 
    \Statex ($M$ and $H$ values in DMH-T and DMH-E)
    \State Feed the aggregation into the prediction network to make final prediction
    \State Return power consumption prediction
  \end{algorithmic}
\end{algorithm}

\begin{table}[t]
\caption{Information of Datasets}
\centering
\begin{tabular}{|l|c|c|c|c|}
\hline
Dataset & Trials & Avg. Records & Training & Testing \\
\hline
AIUT    & 12 &  10,006 &  8 &  4 \\ 
BMW     & 38 &  16,502 & 26 & 12 \\ 
Husky-A & 35 & 157,153 & 23 & 12 \\ 
Husky-B & 35 & 132,197 & 23 & 12 \\ 
Husky-C & 17 &  73,647 &  5 & 12 \\ 
\hline
\end{tabular}
\label{tab:data}
\end{table}
% OK <-
% OK <-

% ====================
\section{Environment Setting}   \label{sec:ES}
% OK ->
We adopt three datasets to evaluate the proposed systems, as described in Section~\ref{sec:ES-data}.
Feature groups for each dataset are then presented in Section~\ref{sec:ES-FG}.
The benchmark systems and neural network settings are listed in Section~\ref{sec:ES-BS}.
% OK <-

\begin{table*}[th]
\caption{Features in each group and dataset}
\centering
\begin{tabular}{|l|l|l|l|l|}
\hline
Dataset & \multicolumn{1}{c|}{$FG_{1}$} & \multicolumn{1}{c|}{$FG_{2}$} & \multicolumn{1}{c|}{$FG_{3}$} \\
\hline
AIUT    & cumulative distance right, cumulative distance left, & lifting plate up-executed, &  rear scanner warning zone active,\\ 
        & left drive activate-manual,  & lifting plate up-in progress,  & lifting plate down-in progress,\\
        & left drive activate-command on, & lifting plate down-executed, & battery cell voltage (3)\\
        & right drive activate-manual,  & frequency of left encoder pulses, & \\
        & right drive activate-command on, & frequency of right encoder pulses (5) & \\

        & lifting plate up-automatic, & & \\
        & lifting plate up-safety interlock, & & \\
        & brake lock-executed, brake release-execute, & & \\
        & automatic mode active, manual mode active, & & \\
        & scanners active zones, scanners safety zones muted, & & \\
        & velocity active zone, fans inactive (16) & & \\
\hline
BMW     & longitudinal acceleration, throttle, motor torque,  & ambient temperature, heating power CAN, & SoC, displayed SoC, AirCon power, \\ 
        & velocity (4) & max battery temperature, elevation, & heat exchanger temperature, \\
        & & battery temperature (5) & regenerative braking signal, \\
        & & & cabin temperature sensor (6)\\
\hline
Husky   & linear\_acceleration\_x, linear\_acceleration\_y, & orientation\_x, orientation\_z, orientation\_w, & position\_x, position\_y,\\ 
        & orientation\_y (3) & velocity\_x, velocity\_y, velocity\_z, & linear\_acceleration\_z, \\
        & & position\_z, temperature, & angular\_x, humidity (5)\\
        & & angular\_y, angular\_z (10) & \\
\hline
\end{tabular}
\label{tab:Fea}
\end{table*}

% OK ->
\subsection{Dataset}   \label{sec:ES-data}
To evaluate the prediction performance, we adopt AIUT, BMW~\cite{BMW}, and Husky~\cite{Husky} datasets.
Table~\ref{tab:data} lists the number of trials (Trials), the average records per trial (Avg. Records), the number of trials for training (Training), and the number of test trials for each dataset (Testing).

AIUT is a real-world AGV dataset collected in 2022 from the smart factory of AIUT Co., Ltd. in Gliwice, Poland.
The AIUT dataset consists of 12 trials with an average of 10,006 records, obtaining various features such as battery, control mode, odometer, board lift, and more.
BMW~\cite{BMW} is provided by the Technical University of Munich in 2020, which consists of 38 running trials of BMW i3 (60Ah) electric vehicles with an average record of 16,502.
The BMW dataset contains features for battery, vehicle, environment, and heating.
Husky~\cite{Husky} is provided by Carnegie Mellon University in 2021, including trials of the Husky A200 unmanned ground vehicle on 4 different routes.
The Husky dataset contains features of battery, motor, location, and environment.
The first 3 routes of the Husky dataset (Husky-A, Husky-B, Husky-C) have 35, 35, and 17 trials of data, respectively.
We ignore route D in Husky because only 5 trials are provided.
Note that the numbers of trials to train and test Husky-C is 5 and 12, designed as few-shot learning tasks for robustness evaluation.
% OK <-

% OK ->
\subsection{Feature Groups}   \label{sec:ES-FG}
In the following experiments, we set the number of the feature groups $H$ to 3, and set the correlation thresholds $T_{1}$ and $T_{2}$ to 0.05 and 0.20, aligned with~\cite{SMC}.
Features in each group and dataset are listed in Table~\ref{tab:Fea}, where feature groups are grouped by Eq.~\ref{eq:FG} and are aligned with~\cite{SMC}.
In Table~\ref{tab:Fea}, values in parentheses are the numbers of features in the corresponding group and dataset.

\begin{table}[th]
\caption{Number of parameters for each system under each dataset and type of network}
\centering
\begin{tabular}{|l|c|c|c|c|c|}
% \hline
% AIUT & BS & FS-S~\cite{SMC} & FS-A~\cite{SMC} & DMH-E & DMH-T \\
% \hline
% MLP         & 84,785 &   X    &   X    &  85,420 &  86,146 \\
% CNN         & 86,537 &   X    &   X    &  86,414 &  87,140 \\
% LSTM        & 87,849 & 85,201 & 86,161 &  84,796 &  85,186 \\
% Transformer & 85,921 &   X    &   X    & 107,548 & 108,010 \\
% \hline
% \hline
% BMW & BS & FS-S~\cite{SMC} & FS-A~\cite{SMC} & DMH-E & DMH-T \\
% \hline
% MLP         & 81,345 &   X    &   X    & 83,404 & 84,457 \\
% CNN         & 82,035 &   X    &   X    & 83,678 & 84,731 \\
% LSTM        & 86,769 & 81,361 & 83,281 & 83,536 & 83,197 \\
% Transformer & 81,281 &   X    &   X    & 89,392 & 83,941 \\
% \hline
% \hline
% Husky & BS & FS-S~\cite{SMC} & FS-A~\cite{SMC} & DMH-E & DMH-T \\
% \hline
% MLP         & 85,185 &   X    &   X    & 87,244 & 88,540 \\
% CNN         & 86,175 &   X    &   X    & 87,686 & 88,982 \\
% LSTM        & 87,129 & 81,361 & 83,281 & 83,956 & 83,860 \\
% Transformer & 86,693 &   X    &   X    & 85,980 & 86,508 \\
% \hline						
\hline									
AIUT	&	MLP	&	CNN	&	LSTM	&	Transformer	\\
\hline									
BS	&	84,785	&	86,537	&	87,849	&	85,921	\\
MVDNN~\cite{MVDNN}  & 85,451  & NA & NA & NA \\
Energy~\cite{ENG} & NA & 86,411  & NA & NA \\
FS-S~\cite{SMC}	&	NA	&	NA	&	85,201	&	NA	\\
FS-A~\cite{SMC}	&	NA	&	NA	&	86,161	&	NA	\\
DMH-E	&	85,420	&	86,414	&	84,796	&	107,548	\\
DMH-T	&	86,146	&	87,140	&	85,186	&	108,010	\\
\hline	
\multicolumn{5}{c}{} \\
\hline								
BMW	&	MLP	&	CNN	&	LSTM	&	Transformer	\\
\hline									
BS	&	81,345	&	82,035	&	86,769	&	81,281	\\
MVDNN~\cite{MVDNN}  & 87,335  & NA & NA & NA \\
Energy~\cite{ENG} & NA & 87,231  & NA & NA \\
FS-S~\cite{SMC}	&	NA	&	NA	&	81,361	&	NA	\\
FS-A~\cite{SMC}	&	NA	&	NA	&	83,281	&	NA	\\
DMH-E	&	83,404	&	83,678	&	83,536	&	89,392	\\
DMH-T	&	84,457	&	84,731	&	83,197	&	83,941	\\
\hline		
\multicolumn{5}{c}{} \\
\hline							
Husky	&	MLP	&	CNN	&	LSTM	&	Transformer	\\
\hline									
BS	&	85,185	&	86,175	&	87,129	&	86,693	\\
MVDNN~\cite{MVDNN}  & 86,707  & NA & NA & NA \\
ENG~\cite{ENG}  & NA & 85,665  & NA & NA \\
FS-S~\cite{SMC}	&	NA	&	NA	&	81,361	&	NA	\\
FS-A~\cite{SMC}	&	NA	&	NA	&	83,281	&	NA	\\
DMH-E	&	87,244	&	87,686	&	83,956	&	85,980	\\
DMH-T	&	88,540	&	88,982	&	83,860	&	86,508	\\
\hline									
\end{tabular}
\label{tab:NPAR}
\end{table}
% OK <-

\subsection{Benchmark Systems}   \label{sec:ES-BS}
% OK ->
In this paper, we employ five benchmark systems, including BS, FS-S, FS-A, ENG, and MVDNN.
BS are the baseline systems with MLP, CNN, LSTM, and transformer networks, similar to the head networks designed in Section~\ref{sec:Sys-DMH-LB} and Table~\ref{tab:ND} with slightly different numbers of fully connected layers and neurons (balance the number of parameters).
In addition, BS is executed on a single machine, and all features are collected on the centralized machine for prediction.
BS is centralized learning rather than distributed learning, adopted to demonstrate the effectiveness of the proposed feature engineering, multi-head, and distributed learning mechanisms.
FS-A and FS-S are LSTM-based systems proposed in~\cite{SMC}, which use all features and selected features (correlation coefficient $\geq$ 0.2, selected in~\cite{SMC}), respectively.
Note that FS-A and FS-S consist only of LSTM networks, without branches of MLP, CNN, and transformer networks.
ENG is a CNN-LSTM-based system proposed in~\cite{ENG}, which extracts spatial and temporal information to predict power consumption.
MVDNN is a multi-view deep neural network proposed in~\cite{MVDNN}, which analyzes the power consumption by multiple views (features).
Note also that all benchmark systems (BS, FS-S, FS-A, ENG, and MVDNN) employ the same features as the proposed DMH-E and DMH-T systems, and have the same number of features and window sizes, but without feature engineering designed in Section~\ref{sec:Sys-FE}.

For a fair comparison, we balance the number of parameters for each system (by the number of neurons in fully connected layers), which are presented in Table~\ref{tab:NPAR}, and NA represents not available.
Note that since there are different numbers of features ($n_{h}$) in each feature group $h$, the same system in different datasets may have different but similar numbers of neurons in layers.
In the transformer-based systems of the AIUT dataset, the DMH-E and DMH-T systems have slightly more parameters than the other systems, due to the higher number of features in $FG_{1}$ (as shown in Table~\ref{tab:Fea}), which lead to larger source and target sequences extractors and more parameters.

All systems are trained with 1,000 epochs, a window size $W$ of 5, a loss function with L1 loss (MAE), an Adam optimizer with a learning rate of 0.1, and a save-best mechanism.
All systems run on a personal computer consisting of an Intel(R) Core(TM) i7-9700 CPU @ 3.00GHZ with 16GB RAM and an NVIDIA GeForce RTX 2080 Ti with 11GB GDDR6.
% OK <-

% ====================
\section{Experimental Results and Discussion}   \label{sec:Exp}
% OK ->
In this section, we first evaluate the proposed systems by predicting power consumption after 1 time step, as shown in Section~\ref{sec:Exp-PE}.
We then perform robustness evaluation by predicting power consumption after 5 and 10 time steps, as described in Section~\ref{sec:Exp-RE}.
The effectiveness of the loss balancing mechanism is demonstrated in Section~\ref{sec:Exp-LB}, and we conduct effectiveness studies in Section~\ref{sec:Exp-CC}.
% OK <-

\begin{table*}[!t]
\caption{Prediction error of power consumption with different head networks after 1 time step}
\centering
\begin{tabular}{|l|rr|rr|rr|rr|rr|}
\hline
& \multicolumn{2}{c|}{AIUT} & \multicolumn{2}{c|}{BMW} & \multicolumn{2}{c|}{Husky-A} & \multicolumn{2}{c|}{Husky-B} & \multicolumn{2}{c|}{Husky-C} \\
\hline
MLP & \multicolumn{1}{c}{MAE} & \multicolumn{1}{c|}{MSE} & \multicolumn{1}{c}{MAE} & \multicolumn{1}{c|}{MSE} & \multicolumn{1}{c}{MAE} & \multicolumn{1}{c|}{MSE} & \multicolumn{1}{c}{MAE} & \multicolumn{1}{c|}{MSE} & \multicolumn{1}{c}{MAE} & \multicolumn{1}{c|}{MSE}\\
\hline
BS & 3.30 E8 & 5.76 E17 & 2,332.94 & 1.99 E7 & 114.28 & 33,278 & 42.54 & 6,826 & 40.68 & 2,748 \\
% FS-A~\cite{SMC} & 725.31 & 7.18 E5 & 9,978.03 & 2.58 E8 & 111.67 & 30,637 & 42.44 & 6,770 & 56.89 & 4,656 \\
% FS-S~\cite{SMC} & 695.92 & 7.01 E5 & 9,978.44 & 2.58 E8 & 111.65 & 30,589 & 42.43 & 6,434 & 57.63 & 4,721 \\
DMH-T & \textbf{610.36} & \textbf{5.57 E5} & 6,058.57 & 1.38 E8 & \textbf{51.35} & \textbf{7,800} & \textbf{29.10} & \textbf{3,107} & \textbf{36.31} & \textbf{2,661} \\
DMH-E & \textbf{636.42} & \textbf{5.93 E5} & 6,350.93 & 1.70 E8 & \textbf{44.57} & \textbf{5,662} & \textbf{24.42} & \textbf{2,254} & \textbf{24.08} & 2,818 \\
\hline
\multicolumn{11}{c}{} \\
\hline
CNN  & \multicolumn{1}{c}{MAE} & \multicolumn{1}{c|}{MSE} & \multicolumn{1}{c}{MAE} & \multicolumn{1}{c|}{MSE} & \multicolumn{1}{c}{MAE} & \multicolumn{1}{c|}{MSE} & \multicolumn{1}{c}{MAE} & \multicolumn{1}{c|}{MSE} & \multicolumn{1}{c}{MAE} & \multicolumn{1}{c|}{MSE}\\
\hline
BS & 654.75 & 6.15 E5 & 9,952.15 & 2.56 E8 & 113.60 & 32,805 & 42.57 & 6,825 & 56.74 & 4,643 \\
% FS-A~\cite{SMC} & 725.31 & 7.18 E5 & 9,978.03 & 2.58 E8 & 111.67 & 30,637 & 42.44 & 6,770 & 56.89 & 4,656 \\
% FS-S~\cite{SMC} & 695.92 & 7.01 E5 & 9,978.44 & 2.58 E8 & 111.65 & 30,589 & 42.43 & 6,434 & 57.63 & 4,721 \\
DMH-T & \textbf{615.91} & \textbf{5.69 E5} & \textbf{6,324.15} & \textbf{1.51 E8} &  \textbf{48.37} &  \textbf{6,589} & \textbf{27.15} & \textbf{3,140} & \textbf{25.03} & \textbf{1,409} \\
DMH-E & \textbf{564.57} & \textbf{4.82 E5} & \textbf{7,272.18} & \textbf{2.09 E8} &  \textbf{48.69} &  \textbf{6,733} & \textbf{19.61} & \textbf{1,921} & \textbf{18.59} &   \textbf{995} \\
\hline
\multicolumn{11}{c}{} \\
\hline
LSTM  & \multicolumn{1}{c}{MAE} & \multicolumn{1}{c|}{MSE} & \multicolumn{1}{c}{MAE} & \multicolumn{1}{c|}{MSE} & \multicolumn{1}{c}{MAE} & \multicolumn{1}{c|}{MSE} & \multicolumn{1}{c}{MAE} & \multicolumn{1}{c|}{MSE} & \multicolumn{1}{c}{MAE} & \multicolumn{1}{c|}{MSE}\\
\hline
BS & 595.10 & 5.32 E5 & 9,941.28 & 2.56 E8 & 112.52 & 28,801 & 42.38 & 6,457 & 65.17 & 5,623 \\
% FS-A~\cite{SMC} & 725.31 & 7.18 E5 & 9,978.03 & 2.58 E8 & 111.67 & 30,637 & 42.44 & 6,770 & \textbf{56.89} & \textbf{4,656} \\
% FS-S~\cite{SMC} & 695.92 & 7.01 E5 & 9,978.44 & 2.58 E8 & \textbf{111.65} & 30,589 & 42.43 & \textbf{6,434} & 57.63 & 4,721 \\
DMH-T & \textbf{589.49} & \textbf{5.22 E5} & \textbf{9,928.98} & \textbf{2.55 E8} & \textbf{111.72} & 29,668 & \textbf{42.25} & 6,569 & \textbf{57.14} & \textbf{4,678} \\
DMH-E & \textbf{589.39} & \textbf{5.22 E5} & \textbf{9,930.96} & \textbf{2.55 E8} & \textbf{111.62} & 29,979 & 42.57 & \textbf{6,387} & \textbf{61.27} & \textbf{5,113} \\
\hline
\multicolumn{11}{c}{} \\
\hline
Transformer & \multicolumn{1}{c}{MAE} & \multicolumn{1}{c|}{MSE} & \multicolumn{1}{c}{MAE} & \multicolumn{1}{c|}{MSE} & \multicolumn{1}{c}{MAE} & \multicolumn{1}{c|}{MSE} & \multicolumn{1}{c}{MAE} & \multicolumn{1}{c|}{MSE} & \multicolumn{1}{c}{MAE} & \multicolumn{1}{c|}{MSE}\\
\hline
BS & 610.89 & 5.54 E5 & 7,558.61 & 1.71 E8 & 126.13 & 39,149 & 42.46 & 6,421 & 60.53 & 5,026 \\
% FS-A~\cite{SMC} & 725.31 & 7.18 E5 & 9,978.03 & 2.58 E8 & 111.67 & 30,637 & 42.44 & 6,770 & 56.89 & 4,656 \\
% FS-S~\cite{SMC} & 695.92 & 7.01 E5 & 9,978.44 & 2.58 E8 & \textbf{111.65} & 30,589 & \textbf{42.43} & 6,434 & 57.63 & 4,721 \\
DMH-T & \textbf{590.42} & \textbf{5.23 E5} & 9,938.80 & 2.56 E8 & \textbf{112.08} & \textbf{29,153} & 43.00 & 6,957 & \textbf{56.51} & \textbf{4,624} \\
DMH-E & \textbf{589.40} & \textbf{5.22 E5} & 8,578.42 & 2.11 E8 &  \textbf{79.44} & \textbf{14,529} & \textbf{39.30} & \textbf{6,092} & \textbf{48.06} & \textbf{3,785} \\
\hline
\end{tabular}
\label{tab:PE01-D}
\end{table*}

\begin{table*}[!t]
\caption{Prediction error of power consumption after 1 time step}
\centering
\begin{tabular}{|l|rr|rr|rr|rr|rr|}
\hline
& \multicolumn{2}{c|}{AIUT} & \multicolumn{2}{c|}{BMW} & \multicolumn{2}{c|}{Husky-A} & \multicolumn{2}{c|}{Husky-B} & \multicolumn{2}{c|}{Husky-C} \\
\hline
Average & \multicolumn{1}{c}{MAE} & \multicolumn{1}{c|}{MSE} & \multicolumn{1}{c}{MAE} & \multicolumn{1}{c|}{MSE} & \multicolumn{1}{c}{MAE} & \multicolumn{1}{c|}{MSE} & \multicolumn{1}{c}{MAE} & \multicolumn{1}{c|}{MSE} & \multicolumn{1}{c}{MAE} & \multicolumn{1}{c|}{MSE}\\
\hline
BS & 8.25 E7 & 1.44 E17 & \textbf{7,446.24} & \textbf{1.76 E8} & 116.63 & 33,508 & 42.49 & 6,632 & 55.78 & 4,510 \\
FS-A~\cite{SMC} & 725.31 & 7.18 E5   & 9,978.03 & 2.58 E8 & 111.67 & 30,637 & 42.44 & 6,770 & 56.89 & 4,656 \\
FS-S~\cite{SMC} & 695.92 & 7.01 E5   & 9,978.44 & 2.58 E8 & 111.65 & 30,589 & 42.43 & 6,434 & 57.63 & 4,721 \\
ENG~\cite{ENG}	&	769.45 	&	7.94 E5	&	\textbf{5,005.30} 	&	\textbf{7.05 E7}	&	90.07 	&	22,955 	&	37.26 	&	\textbf{3,759} 	&	49.50 	&	4,284 	\\
MVDNN~\cite{MVDNN}	&	782.75 	&	8.03 E5	&	7,773.48 	&	1.76 E8	&	113.98 	&	33,078 	&	42.41 	&	6,752 	&	71.44 	&	6,588 	\\
DMH-T & \textbf{601.55} & \textbf{5.43 E5}   & 8,062.62 & 2.00 E8 &  \textbf{80.88} & \textbf{18,302} & \textbf{35.37} & 4,943 & \textbf{43.75} & \textbf{3,343} \\
DMH-E & \textbf{594.94} & \textbf{5.30 E5}   & 8,033.12 & 2.11 E8 &  \textbf{71.08} & \textbf{14,225} & \textbf{31.47} & \textbf{4,163} & \textbf{38.00} & \textbf{3,178} \\
\hline
\end{tabular}
\label{tab:PE01}
\end{table*}

% OK ->
\subsection{Prediction Evalaution}   \label{sec:Exp-PE}
Table~\ref{tab:PE01-D} presents the prediction performances of systems compared with BS, which demonstrate the effectiveness of the proposed feature engineering and DMH.
The values in Table~\ref{tab:PE01-D} are the MAE and mean squared error (MSE) of prediction on test trials, where the bold values outperform the BS.
There are 4 subtables in Table~\ref{tab:PE01-D}, representing performances of MLP-based networks, CNN-based networks, LSTM-based networks, and Transformer-based networks.

On the AIUT dataset, the experimental results in Table~\ref{tab:PE01-D} show that the proposed DMH-T and DMH-E have perfect performance and consistently outperform BS regardless of the used head networks.
On average, DMH-E shows a slight improvement over DMH-T, and significantly outperforms the benchmark system of BS.
On the BMW dataset, the proposed systems outperform BS on CNN and LSTM networks, but are not as good as BS on MLP and Transformer networks.
It is worth mentioning that the BS with MLP network has the best performance on the BMW dataset (with the lowest error), but poorly performed on the AIUT dataset, which shows the instability of the BS model.
On the contrary, the proposed system has stable performance on all datasets.
On Husky datasets (Husky-A, Husky-B, and Husky-C), the proposed DMH-T with MLP, and CNN head networks achieves significant improvements over BS.
DMH-E with MLP, CNN, and Transformer networks show significant improvement over benchmark systems.
The prediction performances of the systems with LSTM networks are close.
Table~\ref{tab:PE01-D} shows that the DMH-T and DMH-E outperform BS on all of the AITU datasets, 8 of 16 of the BMW datasets, and 41 of 48 of the Husky datasets.
% Experimental results in Table~\ref{tab:PE01-D} show that the proposed DMH-T and DMH-E have 100.00\%, 50.00\%, and 85.42\% ($\frac{16}{16}, \frac{8}{16}, \frac{14}{16}$, and $\frac{41}{48}$) probabilities to outperform BS on AITU, BMW, and Husky datasets, respectively.

Table~\ref{tab:PE01} compares the proposed systems to the state-of-the-art systems.
Note that the results of BS, DMH-T, and DMH-E are the averages of performance of MLP-based networks, CNN-based networks, LSTM-based networks, and Transformer-based networks.
The bold values in Table~\ref{tab:PE01} are the top-2 best-performing systems in the column.
Experimental results show that the proposed DMH-T and DMH-E always achieve top-2 performance on the AIUT, HuskyA, and HuskyC datasets, and DMH-E is slightly better than DMH-T.
On the AIUT dataset, the proposed DMH-E reduces the MAE by 18.0\% ($1-\frac{594.94}{725.31}$), 14.5\% ($1-\frac{594.94}{695.92}$), 22.7\% ($1-\frac{594.94}{769.45}$), and 24.0\% ($1-\frac{594.94}{782.75}$) compared with the state-of-the-art systems of FS-A, FS-S, ENG, and MVDNN.
On the HuskyB dataset, the ENG~\cite{ENG} achieves the lowest MSE, but DMH-T and DMH-E still obtain the lowest MAE.
On the BMW dataset, the proposed systems have poor performance as FS-A and FS-S~\cite{SMC}.
We suspect that the large difference in Pearson's correlation coefficients between the training and testing datasets has negative implications for the proposed FS-A and FS-S systems.

As for efficiency, the designed split learning reduces the transmission cost (from client to server) to $\frac{1}{W}$; it only transfers the predicted $FG^{'}_{1, t}, \ldots, FG^{'}_{H, t}$ instead of all features $FG_{h, t-i}, h=1, \ldots, H$ and $i=1, \ldots, W$.
Furthermore, AGV features (local data) are only acquired by the AGV (client).
Neither the prediction system nor other AGVs will know of local data.
The client transmits only the predicted features (power consumption) without the head (local) networks.
Even if the transmission is hacked or snooped (man-in-the-middle attack), the hackers do not know the local network and can not obtain local features through reverse engineering.
The proposed DMH system has a high degree of privacy and security in the smart factory scenario.
In summary, we recommend the DMH-E system, which has the best overall performance and the least amount of information transferred from the client (AGV) to the server (preliminary prediction only).
In addition, the DMH-E system requires less communication bandwidth and reduces the risk of data exposure.
% OK <-

% In each subtable, we compare the performance of 5 systems (in rows of BS, FS-A, FS-S, DMH-E, and DMH-T) on 5 datasets (in columns of AIUT, BMW, Husky-A, Husky-B, and Husky-C), the values in bold are the top-2 best-performing systems in the column.
% Note that FS-A and FS-S consist only of LSTM-based networks, without branches of MLP, CNN, and Transformer networks.
% We directly replicate the results of FS-A and FS-S on LSTM to MLP, CNN, and Transformer.
% Compared with the state-of-the-art systems FS-A and FS-S, the proposed DMH-E significantly reduces MAE by 18.0\% ($1-\frac{594.94}{725.31}$) and 14.5\% ($1-\frac{594.94}{695.92}$), and reduce MSE by 26.2\% ($1-\frac{5.30}{7.18}$) and 24.4\% ($1-\frac{5.30}{7.01}$), respectively.
% On average, the MAE of the proposed DMH-E is 7.3\% ($1-\frac{7446.24}{8033.12}$) higher than that of BS, but is still 19.5\% ($1-\frac{8033.12}{9978.03}$ and $1-\frac{8033.12}{9978.44}$) lower than FS-A and FS-S.
% On average, the proposed DMH-T and DMH-E have top-2 performance and significantly outperform the benchmark systems.
% Compared with the state-of-the-art FS-S, the proposed DMH-E significantly reduces the MAE error by 36.3\% ($1-\frac{71.08}{111.65}$), 25.8\% ($1-\frac {31.47}{42.43}$), and 34.1\% ($1-\frac{38.00}{57.63}$), and reduce the MSE error by 53.5\% ($1-\frac{14225}{30589}$), 35.3 \% ($1 -\frac{4163}{6434}$), and 32.7\% ($1-\frac{3178}{4721}$), respectively on Husky-A, Husky-B, and Husky-C datasets.

\begin{table*}[!t]
\caption{Prediction error of power consumption after 5 time steps}
\centering
\begin{tabular}{|l|rr|rr|rr|rr|rr|}
\hline
& \multicolumn{2}{c|}{AIUT} & \multicolumn{2}{c|}{BMW} & \multicolumn{2}{c|}{Husky-A} & \multicolumn{2}{c|}{Husky-B} & \multicolumn{2}{c|}{Husky-C} \\
\hline
Average & \multicolumn{1}{c}{MAE} & \multicolumn{1}{c|}{MSE} & \multicolumn{1}{c}{MAE} & \multicolumn{1}{c|}{MSE} & \multicolumn{1}{c}{MAE} & \multicolumn{1}{c|}{MSE} & \multicolumn{1}{c}{MAE} & \multicolumn{1}{c|}{MSE} & \multicolumn{1}{c}{MAE} & \multicolumn{1}{c|}{MSE}\\
\hline
BS & 1,320.90 & 7.52 E6 & \textbf{7,899.71} & \textbf{1.81 E8} & 118.92 & 30,201 & 42.09 & 6,662 & 59.41 & 4,923 \\
FS-A~\cite{SMC} & 848.44 & 9.56 E5   & 9,980.60 & 2.58 E8 & 111.59 & 30,365 & 42.58 & 6,835 & 57.42 & 4,701 \\
FS-S~\cite{SMC} & 661.67 & 6.41 E5   & 9,980.45 & 2.58 E8 & 112.41 & 31,803 & 42.83 & 6,916 & 57.38 & 4,698 \\
ENG~\cite{ENG}	&	1,285.20 	&	2.11 E6	&	\textbf{5,217.38} 	&	\textbf{7.72 E7}	&	94.03 	&	\textbf{22,459} 	&	45.11 	&	\textbf{5,309} 	&	50.60 	&	4,032 	\\
MVDNN~\cite{MVDNN}	&	759.94 	&	7.69 E5	&	8,032.62 	&	1.85 E8	&	112.27 	&	28,971 	&	42.58 	&	6,384 	&	62.79 	&	5,301 	\\
DMH-T & \textbf{602.74} & \textbf{5.44 E5}   & 8,375.69 & 2.07 E8 &  \textbf{93.30} & 23,358 & \textbf{40.11} & 6,070 & \textbf{44.54} & \textbf{3,529} \\
DMH-E & \textbf{606.72} & \textbf{5.47 E5}   & 8,229.60 & 1.96 E8 &  \textbf{85.77} & \textbf{19,994} & \textbf{35.21} & \textbf{4,991} & \textbf{45.55} & \textbf{3,672} \\
\hline
\end{tabular}
\label{tab:PE05}
\end{table*}
\begin{table*}[!t]
\caption{Prediction error of power consumption after 10 time steps}
\centering
\begin{tabular}{|l|rr|rr|rr|rr|rr|}
\hline
& \multicolumn{2}{c|}{AIUT} & \multicolumn{2}{c|}{BMW} & \multicolumn{2}{c|}{Husky-A} & \multicolumn{2}{c|}{Husky-B} & \multicolumn{2}{c|}{Husky-C} \\
\hline
Average & \multicolumn{1}{c}{MAE} & \multicolumn{1}{c|}{MSE} & \multicolumn{1}{c}{MAE} & \multicolumn{1}{c|}{MSE} & \multicolumn{1}{c}{MAE} & \multicolumn{1}{c|}{MSE} & \multicolumn{1}{c}{MAE} & \multicolumn{1}{c|}{MSE} & \multicolumn{1}{c}{MAE} & \multicolumn{1}{c|}{MSE}\\
\hline
BS & 1,662.54 & 1.26 E7 & 8,818.13 & 7.35 E8 & 114.01 & 32,114 & 41.88 & 6,598 & 60.14 & 5,001 \\
FS-A~\cite{SMC} & 728.44 & 7.30 E5    & 9,984.16 & 2.58 E8 & 112.17 & 31,547 & 42.26 & 6,520 & 58.03 & 4,758 \\
FS-S~\cite{SMC} & 610.68 & \textbf{5.49 E5}    & 9,984.51 & 2.58 E8 & 111.82 & 31,046 & 42.43 & 6,778 & 56.97 & 4,661 \\
ENG~\cite{ENG}	&	1,434.70 	&	3.00 E6 	&	\textbf{5,717.27} 	&	\textbf{9.36 E7} 	&	106.27 	&	\textbf{21,571} 	&	38.13 	&	\textbf{5,253} 	&	\textbf{51.14} 	&	4,111 	\\
MVDNN~\cite{MVDNN}	&	644.06 	&	6.05 E5 	&	9,847.93 	&	2.24 E8	&	112.11 	&	29,108 	&	42.60 	&	6,378 	&	54.87 	&	4,505 	\\
DMH-T & \textbf{598.65} & \textbf{5.39 E5}    & 8,747.10 & 2.12 E8 & \textbf{100.16} & \textbf{24,523} & \textbf{38.05} & 5,789 & 51.26 & \textbf{4,086} \\
DMH-E & \textbf{607.91} & 5.51 E5    & \textbf{8,391.20} & \textbf{1.97 E8} &  \textbf{98.19} & 25,450 & \textbf{36.60} & \textbf{5,577} & \textbf{48.46} & \textbf{3,989} \\
\hline
\end{tabular}
\label{tab:PE10}
\end{table*}
\begin{table*}[!t]
\caption{Prediction error of DMH-T w/o loss balancing after 5 time steps}
\centering
\begin{tabular}{|l|rr|rr|rr|rr|rr|}
\hline
& \multicolumn{2}{c|}{AIUT} & \multicolumn{2}{c|}{BMW} & \multicolumn{2}{c|}{Husky-A} & \multicolumn{2}{c|}{Husky-B} & \multicolumn{2}{c|}{Husky-C} \\
\hline
Average & \multicolumn{1}{c}{MAE} & \multicolumn{1}{c|}{MSE} & \multicolumn{1}{c}{MAE} & \multicolumn{1}{c|}{MSE} & \multicolumn{1}{c}{MAE} & \multicolumn{1}{c|}{MSE} & \multicolumn{1}{c}{MAE} & \multicolumn{1}{c|}{MSE} & \multicolumn{1}{c}{MAE} & \multicolumn{1}{c|}{MSE}\\
\hline
Without & 602.74 & 5.44 E5 & \textbf{8375.69} & 2.07 E8 & 93.30 & 23,358 & 40.11 & 6,070 & \textbf{44.54} & \textbf{3,529} \\
With    & \textbf{600.99} & \textbf{5.42 E5} & 8505.64 & \textbf{2.05 E8} & \textbf{91.29} & \textbf{21,414} & \textbf{37.78} & \textbf{5,784} & 48.45 & 3,828 \\
\hline
\end{tabular}
\label{tab:LB}
\end{table*}

% OK ->
\subsection{Robustness Evaluation}   \label{sec:Exp-RE}
It is worth noting that smart factories may not provide stable environments, and data collection may be delayed due to various disturbances and interruptions.
Therefore, in this section, we consider the scenarios where data collection is delayed, and power consumption can only be predicted based on features collected in 5 and 10 time steps.
In other words, to evaluate the robustness of the systems, we run the systems to predict the power consumption after 5 and 10 time steps.
Specifically, the system uses the features of $FG_{h, t-i}, h=1, \ldots, H$ and $i=1, \ldots, W$ to predict the power consumption of $P_{t+4}$ and $P_{t+9}$.
Note that the head networks of the proposed DMH-T predicts $FG^{'}_{1, t+4}, \ldots, FG^{'}_{H, t+4}$ and $FG^{'}_{1, t+9}, \ldots, FG^{'}_{H, t+9}$.
Similarly, the head networks of the proposed DMH-E predicts $\text{P}^{'}_{1, t+4}, \ldots, \text{P}^{'}_{H, t+4}$ and $\text{P}^{'}_{1, t+9}, \ldots, \text{P}^{'}_{H, t+9}$.

Table~\ref{tab:PE05} is the power consumption prediction error after 5 time steps, and values of BS, DMH-T, and DMH-E are the average errors of 4 different types of networks (MLP, CNN, LSTM, and Transformer), and values in bold are the top-2 best-performing systems in the column.
Experimental results show that the proposed DMH-T and DMH-E have a similar and the best performance on AIUT and HuskyC datasets.
On the AIUT dataset, the proposed DMH-T reduces the MAE by 29.0\% ($1-\frac{602.74}{848.44}$), 8.9\% ($1-\frac{602.74}{661.67}$), 53.1\% ($1-\frac{602.74}{1285.20}$), and 20.7\% ($1-\frac{602.74}{759.94}$) compared with the state-of-the-art systems of FS-A, FS-S, ENG, and MVDNN.
The proposed systems still obtain good performance on HuskyA and HuskyB datasets, and obtain the top-2 performances in most cases (except the MSEs of DMH-T).
Experimental results of the proposed systems in Table~\ref{tab:PE05} are similar to that of in Table~\ref{tab:PE01}, and demonstrate the stability and robustness of the proposed systems.

Table~\ref{tab:PE10} presents the power consumption prediction error after 10 time steps.
Experimental results show that the proposed DMH-E always achieves the top-2 MAE performance, and the DMH-T achieves the top-2 MAE performance in 3 of the 5 scenarios.
As for MSE measurement, the proposed DMH-T and DMH-E systems achieve the top-2 MSE performance in 3 of the 5 scenarios.
On the AIUT dataset, the proposed DMH-T reduces the MAE by 17.8\% ($1-\frac{598.65}{728.44}$), 2.0\% ($1-\frac{598.65}{610.68}$), 58.3\% ($1-\frac{598.65}{1434.70}$), and 7.1\% ($1-\frac{598.65}{644.06}$) compared with the state-of-the-art systems of FS-A, FS-S, ENG, and MVDNN.

The above experiments demonstrate the robustness of the proposed system, outperforming the state-of-the-art systems by 8.9\% to 53.1\% when predicting 5 time steps after and by 2.0\% to 58.3\% when predicting 10 time steps after.
% OK <-

\begin{figure} 
  \centering
  \begin{subfigure}{1.655in}
    \includegraphics[width=1.655in]{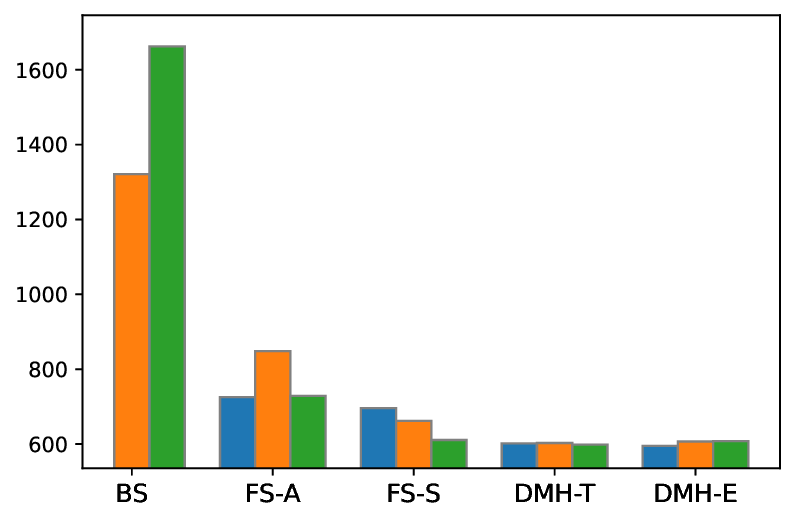}
    \caption{MAE on AIUT dataset}
  \end{subfigure}
  \begin{subfigure}{1.63in}
    \includegraphics[width=1.63in]{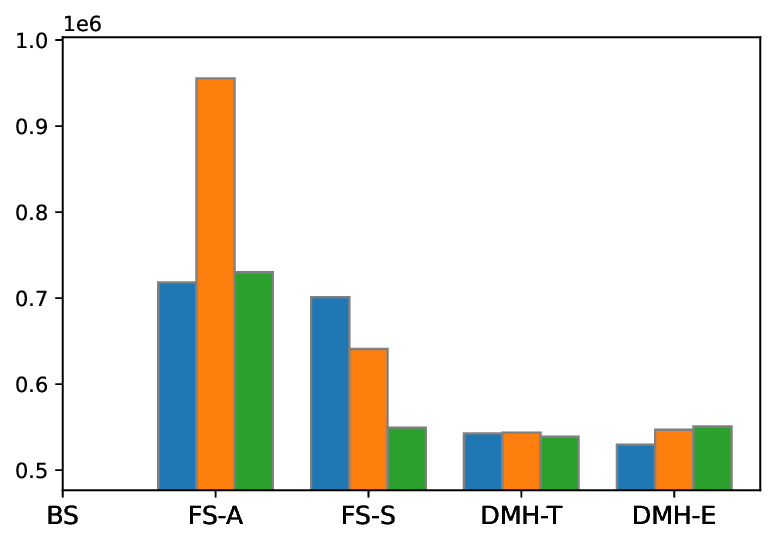}
    \caption{MSE on AIUT dataset}
  \end{subfigure}
  \begin{subfigure}{1.67in}
    \includegraphics[width=1.67in]{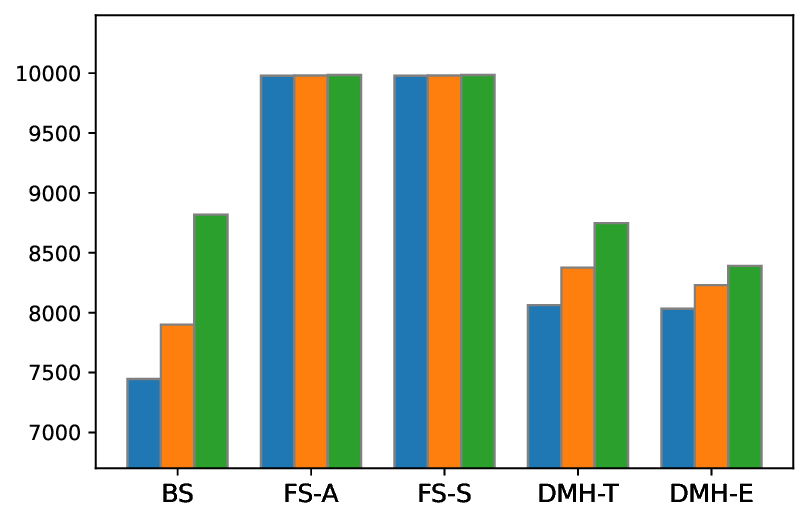}
    \caption{MAE on BMW dataset}
  \end{subfigure}
  \begin{subfigure}{1.63in}
    \includegraphics[width=1.63in]{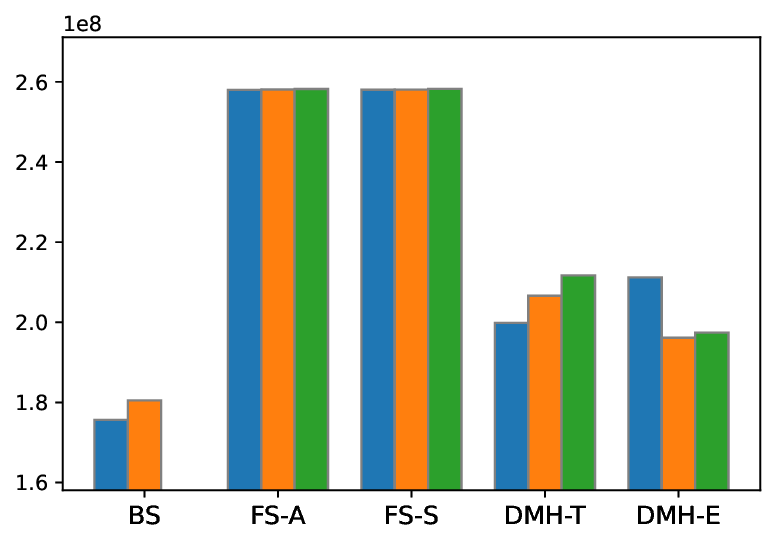}
    \caption{MSE on BMW dataset}
  \end{subfigure}
  \begin{subfigure}{1.58in}
    \includegraphics[width=1.58in]{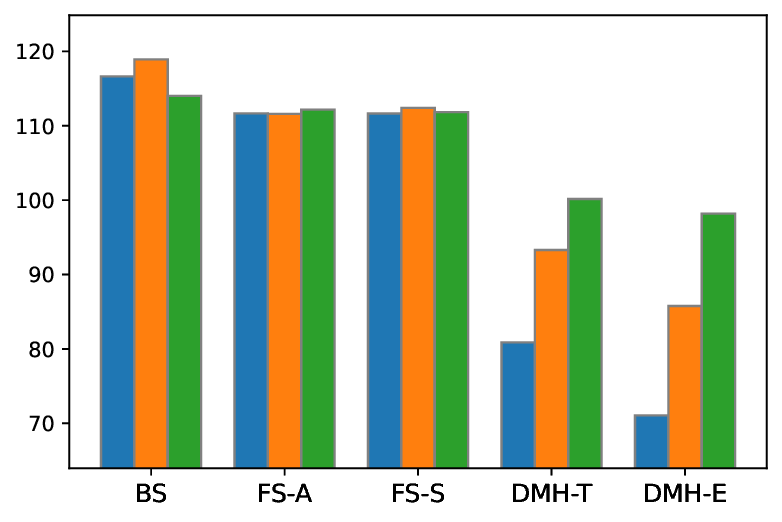}
    \caption{MAE on Husky-A dataset}
  \end{subfigure}
  \begin{subfigure}{1.63in}
    \includegraphics[width=1.63in]{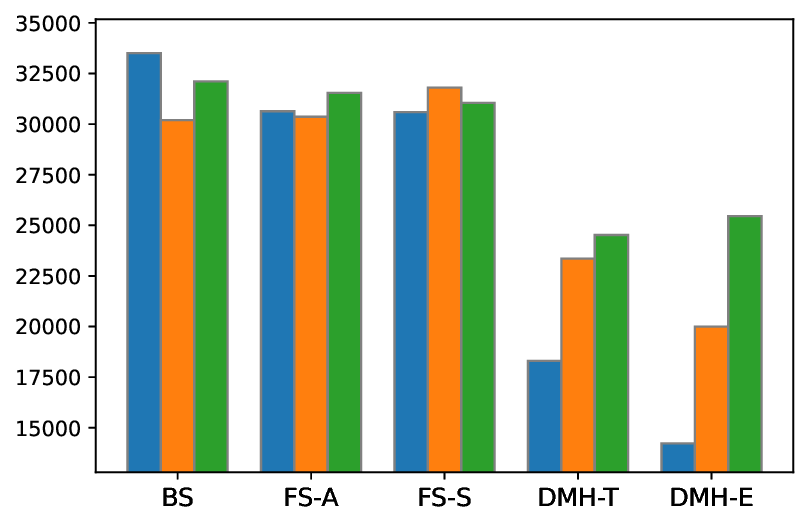}
    \caption{MSE on Husky-A dataset}
  \end{subfigure}
  \begin{subfigure}{1.58in}
    \includegraphics[width=1.58in]{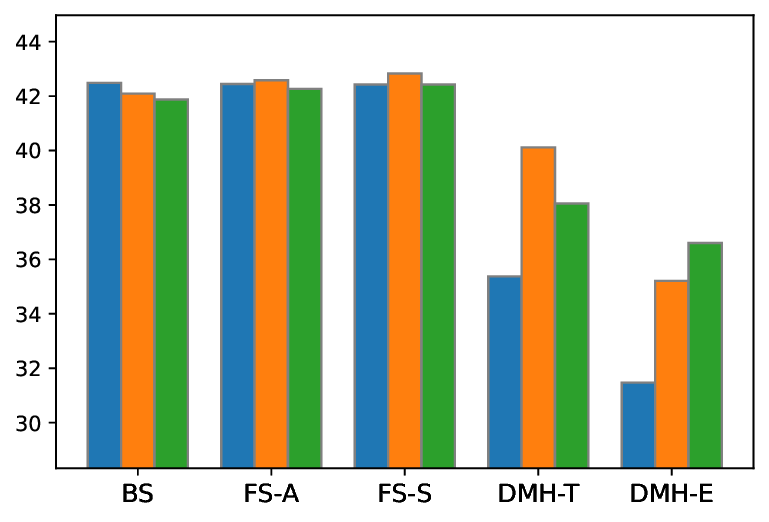}
    \caption{MAE on Husky-B dataset}
  \end{subfigure}
  \begin{subfigure}{1.63in}
    \includegraphics[width=1.63in]{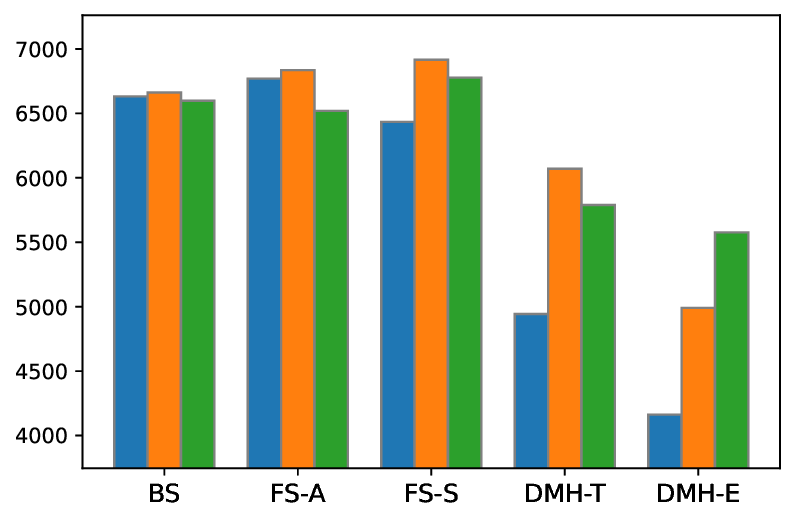}
    \caption{MSE on Husky-B dataset}
  \end{subfigure}
  \begin{subfigure}{1.58in}
    \includegraphics[width=1.58in]{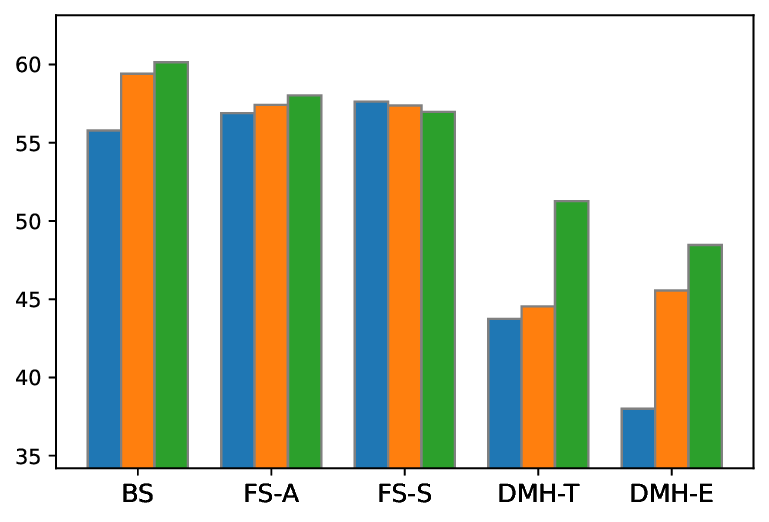}
    \caption{MAE on Husky-C dataset}
  \end{subfigure}
  \begin{subfigure}{1.63in}
    \includegraphics[width=1.63in]{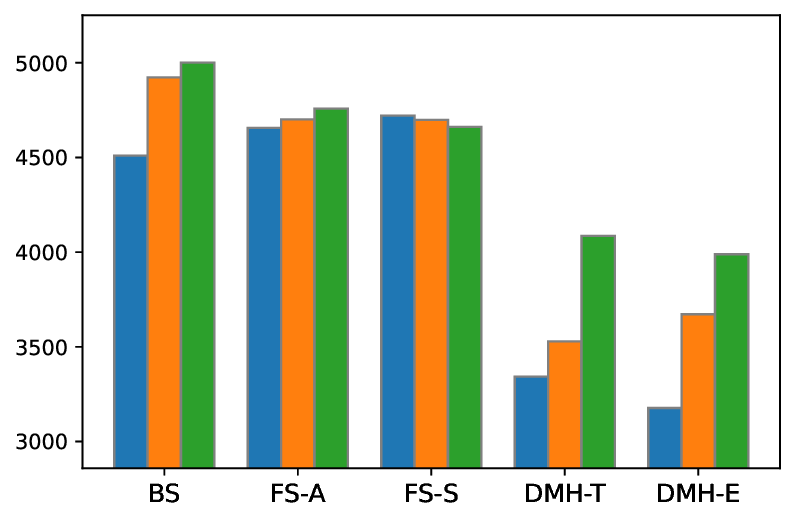}
    \caption{MSE on Husky-C dataset}
  \end{subfigure}
  \caption{Effectiveness of modules (blue, orange, and green bars are for 1, 5, and 10 time steps, respectively)}
  \label{fig:CC}
\end{figure}

% OK ->
\subsection{Effectiveness of Loss Balancing Mechanism}   \label{sec:Exp-LB}
In this section, we investigate the effectiveness of the loss balancing mechanism on DMH-T, as shown in Table~\ref{tab:LB}, which lists the results for DMH-T with and without loss balancing.
On the AIUT, Husky-A, and Husky-B datasets, both MAE and MSE decrease when loss balancing is applied.
However, on the Husky-C dataset, both MAE and MSE increase when loss balancing is applied.
In summary, there is a high probability that the loss balancing mechanism improves the DMH-T predictions with the largest MAE improvement of 5.8\% ($1-\frac{37.78}{40.11}$) on the Husky-B dataset.
% OK <-

% OK ->
\subsection{Effectiveness of the Modules}   \label{sec:Exp-CC}
In this section, we study the effectiveness of the modules.
BS system has traditional deep neural networks (DNN).
FS-S system has DNN and feature engineering.
The proposed DMH systems have DNN, feature engineering, and multi-head learning.
The performance comparison between BS, FS-S, and DMH represents the effectiveness of the feature engineering and multi-head learning mechanisms, as shown in Figure~\ref{fig:CC}.

In Fig.~\ref{fig:CC}, the blue, orange, and green curves are the average losses (MLP, CNN, LSTM, and Transformer) of predicting power consumption after 1, 5, and 10 time steps, respectively.
The x-axis represents the systems of BS, FS-A~\cite{SMC}, FS-S~\cite{SMC}, DMH-T, and DMH-E, and the y-axis represents the MAE or MSE loss.
Note that some losses in BS are extremely large that we ignore them to clearly demonstrate the trends.

From Fig.~\ref{fig:CC}, it can be found that the blue bars are usually the lowest, which is consistent with common sense that the prediction of 1 step after is more accurate than 5 or 10 steps after.
Furthermore, regarding the losses of the AIUT and Husky datasets, the systems on the right side have lower prediction losses than the systems on the left side.
The phenomenon first demonstrates the effectiveness of feature engineering based on Pearson's correlation coefficient.
Feature selection of FS~\cite{SMC} can improve the prediction of BS.
The results also demonstrate that feature grouping based on correlation and the proposed multi-head learning mechanisms further enhance prediction (comparing FS and DMH systems).
% OK <-

% OK ->
% ====================
\section{Conclusion}   \label{sec:Con}
As more and more automatic vehicles, power consumption prediction becomes a vital issue for charging scheduling, task scheduling, and energy management.
Most research focuses on automatic vehicles in transportation, but little attention has been paid to automatic ground vehicles (AGVs) in smart factories, which face complex environments and generate large amounts of data.
There is an inevitable trade-off between feature diversity and interference; therefore, how to organize the collected features is a critical issue for power consumption prediction.
In this paper, we propose Distributed Multi-Head learning (DMH) systems for power consumption prediction in smart factories.
DMH systems group features according to Pearson's correlation.
Multi-head learning mechanisms are then proposed to reduce noise interference.
The head network is designed to predict the corresponding features at the next time step, which are aggregated to predict power consumption.
Additionally, DMH systems are designed for distributed learning, the head networks are held by the client, and the prediction network is held by the server.
DMH systems reduce the client-to-server transmission cost, and share knowledge without sharing local data and the entire model to enhance the privacy and security levels.
Experimental results show that the proposed DMH systems rank in the top-2 on most datasets and scenarios.
DMH-E system reduces the MAE of the state-of-the-art systems by 14.5\% to 24.0\%.
For robustness evaluation, the proposed DMH-E still outperforms the state-of-the-art system by 8.9\% to 53.1\% (after 5 time steps) and 2.0\% to 58.3\% (after 10 time steps).
Effectiveness studies demonstrate the effectiveness of Pearson's correlation-based feature engineering, and feature grouping with the proposed multi-head learning can further enhance prediction performance.
In conclusion, we recommend the DMH-E system that has the lowest overall error, the lowest communication bandwidth requirements, and the lowest risk of data exposure. In the future, we will break through the limitations of current constraints, such as extremely low client processing power, transmission errors, and delays, and also focus on heterogeneous knowledge transfer.
% OK <-

\section*{Acknowledgments}
This work is partially supported by the National Centre for Research and Development under the project Automated Guided Vehicles Integrated with Collaborative Robots for Smart Industry Perspective, and the Project Contract no. is: NOR/POLNOR/CoBotAGV/0027/2019 -00.

% {\appendix[Proof of the Zonklar Equations]
% Use $\backslash${\tt{appendix}} if you have a single appendix:
% Do not use $\backslash${\tt{section}} anymore after $\backslash${\tt{appendix}}, only $\backslash${\tt{section*}}.
% If you have multiple appendixes use $\backslash${\tt{appendices}} then use $\backslash${\tt{section}} to start each appendix.
% You must declare a $\backslash${\tt{section}} before using any $\backslash${\tt{subsection}} or using $\backslash${\tt{label}} ($\backslash${\tt{appendices}} by itself
%  starts a section numbered zero.)}


\bibliographystyle{IEEEtran}
\bibliography{refs}

\vfill
\end{document}